\documentclass{article}

\usepackage{arxiv}

\usepackage[utf8]{inputenc} 
\usepackage[T1]{fontenc}    
\usepackage{hyperref}       
\usepackage{url}            
\usepackage{booktabs}       
\usepackage{nicefrac}       
\usepackage{microtype}      
\usepackage{lipsum}		    
\usepackage{graphicx}
\usepackage{doi}
\usepackage{multicol}
\usepackage[bottom]{footmisc}
\usepackage{amsfonts, amsmath, times, url}
\usepackage{amssymb}
\usepackage{subfigure}
\usepackage{color}
\usepackage{algorithmic, algorithm}

\title{On the use of Wasserstein metric in topological clustering of distributional data}

\date{}	

\author{Gu\'ena\"el Cabanes \\
    LIPN-CNRS, UMR 7030\\
    Universit\'e Sorbonne Paris Nord\\
    99 Avenue J-B. Cl\'ement, 93430 \\ Villetaneuse, France\\
    \texttt{cabanes@lipn.univ-paris13.fr} 
	\And
	Youn\`es Bennani \\
    LIPN-CNRS, UMR 7030\\
    Universit\'e Sorbonne Paris Nord\\
    99 Avenue J-B. Cl\'ement, 93430 \\ Villetaneuse, France\\
    \texttt{younes@lipn.univ-paris13.fr}     
	\And
	Rosanna Verde \\
    Dip. Matematica e Fisica\\
    Universit\'a della Campania "\textit{Luigi Vanvitelli}" \\ 
    Viale A. Lincoln, 5, 81100 \\
    Caserta,  Italy
    \And
	Antonio Irpino \\
    Dip. Matematica e Fisica\\
    Universit\'a della Campania "\textit{Luigi Vanvitelli}" \\ 
    Viale A. Lincoln, 5, 81100 \\
    Caserta,  Italy
}

\renewcommand{\shorttitle}{Wasserstein metric in Topological Clustering}

\hypersetup{
pdftitle={Wasserstein metric in Topological Clustering},
pdfsubject={cs.ML},
pdfauthor={Gu\'ena\"el Cabanes},
pdfkeywords={Clustering, Self-Organising Map, Histogram data, Wasserstein distance, Density measure},
}

\begin{document}
\maketitle

\begin{abstract}
	This paper deals with a clustering algorithm for histogram data based on a Self-Organizing Map (SOM) learning. It combines a dimension reduction by SOM and the clustering of the data in a reduced space. Related to the kind of data, a suitable dissimilarity measure between distributions is introduced: the $L_2$ Wasserstein distance. Moreover, the number of clusters is not fixed in advance but it is automatically found according to a local data density estimation in the original space. Applications on synthetic and real data sets corroborate the proposed strategy.
\end{abstract}


\section{Introduction}
Nowadays, the Big Data era is characterized by huge amounts of data to be managed and analyzed and histogram data represents a useful tool for summarizing sequences of numerical data. Some examples are financial data in economics applications, sensor data for environmental phenomena detection, or energy consumption data loaded by smart meters. Further sources of histogram data arise from Official Statistics Institutes, that make available results of surveys only in form of aggregated or summarized data for preserving the privacy of respondents. A similar case occurs when repeated data are observed on individuals, for example, from a bank or an hospital. The main advantage of histogram data is in the possibility to take into account the shape of the distribution of the data with a smaller loss of information than a synthesis of the observed data by means of the position and scale indexes (for example, by arithmetic mean and standard deviation, only).

In Symbolic Data Analysis (SDA) framework a \textit{histogram variable} has been defined by Bock and Diday \cite{Bock00} as a multi-valued variable, assuming that each individual is described by an estimated probability measure, like a histogram. According to the given definition, a histogram
is expressed by a sequence of disjoint intervals with weights(e.g., relative frequencies) associated. Recently, several techniques have been developed for this kind of data, referred to distributional data analysis \cite{IrVer06,Arr09,IrpVer2015,IrpVer2015reg,Brito15}.

In the field of research on distributional data analysis, among the different metrics to compare distributions \cite{Gibbs}, the $L_2$ Wasserstein distance (see \cite{Rush01}) has been chosen for the interesting interpretative properties that it presents, some of them have recalled or verified in \cite{IrpVer2015reg}.

 Irpino et al. in \cite{IrpinoR07} proved that the $L_2$ Wasserstein distance between two histograms can be computed through the distances between the centers and one third of the distances between radii of the intervals (bins).
 Another interesting decomposition of the $L_2$ Wasserstein distance, discussed in \cite{IrpVer2015}, consists in a decomposition into two components: a first related to the means (location parameter) and a second related to the scale and shape parameters of the compared distributions. In such a way, the results of distributional data analysis take into account, separately, the role in the analysis of the differences in position, scale and shape  of the data. Here, we focus especially on the first type of decomposition, that is in the distances between centers and radii, to simplify computational aspects and, in the same time, use a consistent distance measure to compute the average histograms based on the Fr\'echet mean definition.

In this paper, we present a two-level clustering method for histogram data based on a Self-Organizing Map learning. The Self Organizing Map (SOM) \cite{Kohonen01} is an unsupervised neural network algorithm that maps high-dimensional input data vector in a reduced subspace through a competitive learning strategy. Self Organizing Map uses a neighborhood function to preserve the topology of original input data space in a reduced subspace.

 SOM is also a popular non-linear technique for unsupervised learning and data visualization. The learning of a Self-Organizing Map is proposed as an efficient method to address the problem of clustering, especially for high-dimensional data having an input topological structure. Once a (usually, rectangular or hexagonal) grid of nodes is chosen for defining the topology of the map, the procedure of mapping a vector from data space onto the map consists in finding the node with the closest (smallest distance metric) weight vector to the data space vector.

The development of a SOM method to histogram data is suitable to analyze data that are already available in aggregated form (like some Official Statistics or confidential data) or they are generated as syntheses of huge amount of original data.
The histograms, as empirical distributions, are able to preserve much more information of data than simple means and standard deviations that are usually used to synthesize data-set.

SOM for symbolic data was introduced by Bock \cite{Bock00} in 2000, to clustering and  visualize in a reduced subspace the data endowed of a topological structure. Other SOM methods, developed especially for interval data, have been proposed by using adaptive distances \cite{IRVEPAT08,Carvalho16} or based on relational matrices \cite{Carvalho16}. In the analysis of histogram data, a batch version of SOM has been proposed by \cite{COMPSTAT2012} based on the $L_2$ Wasserstein distance. Adaptive Wasserstein distances has been also developed in this context, to find, automatically, weights for the variables for the whole data-set as well as for each node. However, these methods can provide a quantification and a visualization of symbolic data (intervals, histograms) but they are not able to partition the data in a set of homogeneous clusters. A two-level clustering algorithm based on SOM, DS2L-SOM learning for interval data has been also proposed by \cite{Cabanes13}, that combines the dimension reduction, achieved by a SOM learning step, and groups the data in a reduced space in a certain number of homogeneous clusters.

Here, we propose an SOM-based two-level clustering on histogram data. In the partitioning phase is used the $L_2$ Wasserstein distance, as proposed in dynamic clustering algorithm for histogram data (see \cite{IRVERLEC06}). An interesting advantage of this approach is that the number of cluster is not set in advance among the parameters of the clustering algorithm. It is automatically induced by a criterion related to the estimation of a local density of the data in the original space.
Thanks to the abstraction power of the histogram representation (as a summary of punctual input data) and the linear complexity of the proposed algorithm, the  approach is adapted to the analysis of large-dimensional data-sets. Tests on simulated data have been performed for showing the capability of the method in preserving the information about the distributions describing each data. An application on real data shows the usefulness the method.

The rest of the paper is organized as follows. Section \ref{L2} recalls the definition of  histogram data and the $L_2$ Wasserstein distance. In section \ref{S2L} we present the proposed algorithm. Section \ref{Experiments} shows the experimental protocol and the results obtained to validate our approach. In section \ref{Application} an application on a climatic data-set is proposed. Finally, section \ref{Conlu} concludes the paper also giving some prospects.

 \section{Histogram data and the $L_2$ Wasserstein distance}
 \label{L2}
Let's denote $X$ a histogram variable. The learning sample $x=\{x^{k}\}$ (for $k=1, \ldots,N$) is a set of $N$ histogram data.
 \\
 $x^{k}$ is an estimated distribution. That is defined by a sequence of continuous and no-overlapped intervals (or bins) $I^k_v$ with associated weights or relative frequencies $\pi_v$  (for $v=1, \ldots, h$), such that $\sum_{v=1}^h \pi_{v}=1$:
 \begin{equation}
 x^{k}=[(I^k_{1},\pi_{1}), \ldots,(I^k_{v},\pi_{v}),\ldots,(I^k_{h},\pi_{h})]
 \end{equation}

 Assuming a uniform density for each $I_{v}$, an empirical distribution function $F(x)$ is associated to each histogram, that is, a cumulated relative frequency function. Its inverse $F^{-1}(t)$ is the quantile function, denoted hereafter $Q(s)$, that is, a piecewise linear function with domain in $[0,1]$.

To compare histogram data, we use the $L_2$ Wasserstein metric \cite{Rush01} (also named Mallow's distance \cite{Mall72}).
In the case of distributions defined on $\mathfrak{R}$ \cite{Pana16}, this metric is defined as follows:

 \begin{equation}
     d^2_W (x^{k} ,x^{l)} ): = {\int \limits_{0}^1 {\left( {Q_{k}(s) - Q_{l}(s)} \right)^2 ds} }.
 \end{equation}

 where $Q_{k}(s)$ and $Q_{l}(s)$ are the quantile functions associated with the $x^{k}$ and $x^{l}$ histogram data.

  Since the quantile functions are piecewise linear, the integral is not solved globally but trough the sum of simple integrals defined on the bounds of each pair of corresponding pieces of the quantile functions. That is:

 \begin{equation}
     d^2_W (x^{k} ,x^{l)} ): = \sum\limits_{i=1}^h {\int\limits_{q_{v-1}}^{q_v} {\left( {Q_{k}(s) - Q_{l}(s)} \right)^2ds} }.
 \end{equation}
 
 where: $q_v= \sum_{i=1}^v \pi_i$ (for $i=1, \dots, h$) are cumulated relative frequencies.
 
 That requires the all histogram data have to be homogenized (in a preprocessing step) in order to be compared on same set of quantile values $q_v$ of the quantile functions. To overcome this lack, it is easier to take equi-frequency or equi-depth histograms.

The values of the bins are assumed uniformly distributed, so that, each interval $I_v$ can be represented through its center and radius, as follows: $c+r(2t-1)$ for $0 \le t \le 1$. According to this expression, Irpino et al. \cite{IrVer06} proved that the $L_2$ Wasserstein distance
 $d^2_W (x^{k} ,x^{l})$ can be rewritten in terms of centers and radii of bins of the two histogram data, by:

 \begin{equation}\label{HOMSQfin}
     d^2_W (x^{k} ,x^{l}): =  \sum\limits_{v=1}^h {\pi_{v} \left[{\left({{c^k_{v}-c^l_{v}}}\right)^2 + \frac{1}{3} \left({{r^k_{v}-r^l_{v}}}\right)^2}\right]}.
 \end{equation}

 This definition simplifies the computational procedure and it gives an interesting interpretation of the $L_2$ Wasserstein distance in terms of weighted sum of the squared Euclidean distances between the centers and the radii of the bins of the histograms.

Based on this expression, \cite{IRVERLEC06} have shown that the "average histogram" can be obtained by the means of the centers and the means of the radii of the bins of the $N$ histograms. The "average", or "barycenter histogram", denoted $x^b$, is expressed as follows:

 \begin{equation}\label{bar}
  \bar{x}=\{([\bar{c}_{v}-\bar{r}_{v};\bar{c}_{v}+\bar{r}_{v}],\pi_v)\}_{v=1,\ldots,h}
 \end{equation}

 where:
 \begin{equation}
  \bar{c}_{v}  = N^{-1}\sum\limits_{i = 1}^N { c^i_{v} }  \hspace{10pt};\hspace{10pt}
  \bar{r}_{v}  = N^{-1}\sum\limits_{i = 1}^N { r^i_{v} }  \hspace{10pt}(v=1,\ldots,h).
 \end{equation}

The component of the $L_2$ Wasserstein distance, related to the distances between the centers of the bins, allows of interpreting the differences between the histograms according to the location parameters, taking into account the variability of the centers; while the component related to the distances between the radii of the bins, highlights the differences in variability and shape of the the distributions.

 Moreover, a measure of variability (mean sum of squares) for the $N$ histogram data can be expressed by the mean of the $L_2$ Wasserstein distances $d^2_W(.)$ between the histogram data $x^k$ (for $k=1, \ldots, _N$) and the average histogram $x^b$, that is:
 \begin{equation}
 TI=\sum\limits_{k=1}^{N}{d^2_W(x^k,\vec{x})}.
 \end{equation}

The recalled properties of the $L_2$ Wasserstein distance, allow of generalizing the concept of mean, of the Fr\'echet mean, and of the variance or inertia of a set of histogram data. Those results are also useful in the context of Clustering of a set of histogram data, in a certain number of disjointed clusters. In \cite{IRVERLEC06} is shown that it is possible to keep the classical internal validation indexes, based on the decomposition of the Total Sum of Squares into the Within Sum of Squares and Between Sum of Squares of the clusters, according to the Huygens' theorem.

\section{DHSOM: a topological density-based clustering for Histogram data}
\label{S2L}
We propose here a method to learn the structure of histogram data, based on the automated enrichment and segmentation of a group of prototypes computed by a modified version of the Self-Organizing Map (SOM) adapted to histograms \cite{COMPSTAT2012}. This method extends the DS2L-SOM algorithm, proposed in \cite{Cabanes12}, to histogram data.
\subsection{Principles of the approach}
A SOM consists of a set of artificial neurons that represent the data structure. Neurons are connected with their neighbors according to topological connections (also called neighborhood connections). The input data-set is used to organize the SOM under topological constraints of the input space. Thus, a correspondence between the input space and the mapping space is built such that, two close observations in the input space, should activate the same neuron, or two neighboring neurons, of the SOM. A prototype describes each neuron and, to respect the topological constraints, neighboring neurons of the Best Matching Unit of a data (BMU, the most representative neuron) also update their prototypes for a better representation of this data. This update is important because the neurons are close neighbors of the best neuron.

In DS2L-SOM \cite{Cabanes12}, for providing an estimation of the underlying distribution of the data, prototypes are enriched with local density and connectivity estimations. More specifically, using a Gaussian kernel estimator \cite{Silverman81}, we estimate a local density as a measure of the data density surrounding the prototype. The connectivity measures how close are to prototypes for the data representation. The connectivity value of a pair of prototypes is the number of data that are well represented by both of them (the two prototypes are the first two Best Match Units for these data).
From this estimation, it is possible to cluster the prototypes (as a representation of the data partition) as described in \cite{Cabanes12}. In that case, clusters are defined as regions of the representation space having a relative high density and  separated by regions of relative low density. As in most density-based methods, the number of clusters is detected automatically.

To adapt the principles of DS2L-SOM to histogram data, we need a modified version of the Self-Organizing Map and an adapted enrichment of the prototypes. We chose here a SOM algorithm for histogram data that have been proposed in \cite{COMPSTAT2012}, where each prototype is defined as a histogram and the distances between data and prototype are computed with the $L_2$ Wasserstein distance. In addition, the estimation of the local densities and variabilities in DS2L-SOM are mainly based on the distance between the data and the prototype. By using the $L_2$ Wasserstein distance in the enrichment step, the clustering of histogram data becomes possible.

\subsection{SOM for histogram data}
The adaptation of SOM to histogram data is based on two principles: each prototype is a histogram and the distances between observations and prototypes are computed with the $L_2$ Wasserstein metric. In this paper we propose the use of a batch version of SOM adapted to histograms.

The fist phase of the algorithm is the \textit{Competition step}, where each observation is assigned to the neuron according to the closest prototype (i.e. the $BMU$) according to the $L_2$ Wasserstein metric. The second phase is the \textit{Adaptation step}, where each prototype is updated such that the average Wasserstein distance between all the prototypes and the observations is minimized consistently with the topological structure of the map.

The function to minimize is the following:
\begin{equation}\label{cost}
{R}(w) = \sum_{k=1}^N \sum_{i=1}^M K_{i u^*(x^{k})} d^2_W(w^i,x^{k})
\end{equation}
where $x$ is an observation represented as a histogram, $w$ is a prototype (an histogram representing a set of observations), $N$ is the size of the learning data-set, $M$ represents the number of neurons in the map, $u^*(x^{k})$ is the neuron having the weight vector closest to the observation $x^{k}$ (namely, the $BMU$), and $K_{ij}$ is the neighborhood function: a positive symmetric kernel function. The relative importance of a neuron $i$ with respect to a neuron $j$ is weighted by the value of the kernel function $K_{ij}$ which is defined as:
\begin{equation*}
	K_{i,j} = \frac {1}{\lambda (t)} \times e^{-\frac {d^2_1(i,j)}{\lambda^2 (t)}}.
\end{equation*}
The $\lambda(t)$ term is the \textit{temperature function} that models the topological neighborhood extent. It is defined as:
\begin{equation*}
  \lambda ( t )  = \lambda_i ( \frac{\lambda_f} {\lambda_i} )^{  \frac{t}{t_{max}}}
\end{equation*}
where $\lambda_i$ and $\lambda_f$ are the initial and the final temperature, and $t_{max}$ is a parameter representing the number of iterations. The $d_1(i,j)$ term is the Manhattan distance defined between two neurons $i$ and $j$ on the map grid, having coordinates $(k,m)$ and $(r,s)$ respectively:
\begin{equation*}
 d_1 (i,j) = \parallel r - k \parallel + \parallel s - m \parallel.
\end{equation*}
To minimize eq. (\ref{cost}), each prototype is updated to approximate the barycenter of the observations, weighted by $K_{ij}$. Prototypes being computed using eq. (\ref{bar}), the weighted barycenters are expressed as follow:
 \begin{equation} \label{adapt}
  \bar{w}^j=\{([\bar{c}^j_{v}-\bar{r}^j_{v};\bar{c}^j_{v}+\bar{r}^j_{v}],\pi^j_v)\}_{v=1,\ldots,h}
 \end{equation}
 where:
 \begin{equation} \label{adapt2}
  \bar{c}^j_{v}  = \frac{\sum\limits_{i = 1}^N {K_{ij} c^i_{v} }}{\sum\limits_{i = 1}^N {K_{ij}}}
 \end{equation}
 and
 \begin{equation} \label{adapt3}
  \bar{r}^j_{v}  = \frac{\sum\limits_{i = 1}^N {K_{ij} r^i_{v} }}{\sum\limits_{i = 1}^N {K_{ij}}}.
 \end{equation}
We remark that the $L_2$ Wasserstein distance between histograms can be interpreted as an Euclidean distance between the respective quantile functions, and that a linear combination of quantile function is again a quantile function only if the weights are positive. Since a  one-to-one correspondence exists between a histogram and its quantile function, it is equivalent to consider the prototype as a histogram even if its quantile function is updated in the algorithm. The complete algorithm is described in algorithm \ref{SOMhist}.

\begin{algorithm}[!ht]
\caption{SOM for histogram data}
\label{SOMhist}
\begin{algorithmic}[1]
    \STATE{Define the topology of the SOM.}
    \STATE{Initialize the prototypes $w^{j}$.}
    \REPEAT
      \FORALL {histogram data $x^{k}$} 
          \STATE{Among the $M$ prototypes, choose $u^{*}(x^{k})$ according to the $L_2$ Wasserstein distance $d^2_W$:
 	      \[u^{*}(x^{k}) =  \mathop{Argmin}_{1 \leq i\leq M} d_W^2(x^{k}, w^{i})\]}
      \ENDFOR
       \FORALL {prototype $w^{i}$} 
          \STATE{Update $w^{i}$ according to eq. \ref{adapt}, \ref{adapt2} and \ref{adapt3}.}
      \ENDFOR
 \UNTIL{$t = t_{tmax}$}
\end{algorithmic}
\end{algorithm}

\subsection{Prototypes Enrichment}
To improve the representation of the underlying structure of the data once the prototypes are computed, the model can be enriched with additional information associated with each prototype  (algorithm \ref{enrichalgo}). Two additional information are computed in this step: the connectivity and the local density.

The local density $D_{i}$, associated with each prototype $w^i$, is estimated as follow:
 \begin{equation}
 D_{i} = 1/N \sum_{k=1}^{N} \frac{1}{ \sigma \sqrt{2\pi}} e^{-{\frac{d^2_W(w^{i},x^{k})} {2 \sigma^2}}}
 \end{equation}
where $\sigma$ is a bandwidth parameter chosen by the user and $d^2_W(w^i,x^{k)}$ are the $L_2$ Wasserstein distances between the $M$ prototypes $w^i$ and the $N$ histogram data $x^k$. The proposed method for estimating the mode density follows the proposal of \cite{Pamudurthy07}, with an adaptation to histogram data. It has been shown that when the number of data approaches infinity, the estimator $D$ converges asymptotically to the true density function \cite{Silverman86}. It is worth noting that the selection of the $\sigma$ parameter affects the final results. If $\sigma$ is too large, all data will influence the local density of the representation space around the prototypes. It follows that close prototypes will be associated with similar densities and the accuracy of the estimate will decrease. On the other hand, if $\sigma$ is too small, a large proportion of data (the most distant prototypes) will not influence the density of the prototypes, inducing a loss of information. A heuristic that seems capable to  give good results suggests to define $\sigma$ as the average distance between a prototype and its nearest neighboring one.

The connectivity between neurons is a measure of discontinuity in the topological space. It allows to detect clusters that are separated by empty regions in the representation space. Since such regions are not often well represented by the prototypes, the connectivity measure assure the detection of cluster borders between two adjacent neurons. However, if a boundary is defined between a region of lower density and two regions with a higher density, the connectivity may not be sufficient to detect borders and an estimation of local densities is necessary.

\begin{algorithm}
\caption{Prototypes enrichment}
\label{enrichalgo}
\begin{algorithmic}
 \REQUIRE The Wasserstein distance between each observation and each prototype
  \ENSURE A density value $D_{i}$ for each neuron $w^i$ and a connectivity value $v_{i,j}$ for each pair of neurons $i$ and $j$.
   \FORALL {neuron $i$}
    \STATE{Compute the local density $D_i$ using: \[D_{i} = 1/N \sum_{k=1}^{N} \frac{e^{-{\frac{d^2_W(w^{i},x^{k})} {2 \sigma^2}}}}{ \sigma \sqrt{2\pi}} \] with $\sigma$ a bandwidth parameter chosen by user.}
     \ENDFOR
      \FORALL {data $x^k$}
          \STATE{Find the two closest prototypes (BMUs) $u^*(x^k)$ and $u^{**}(x^k)$ using:
		\[u^*(x^k) = argmin_i d^2_W(w^i,x^{k})\]}
          \STATE{Compute $v_{i,j} = $ the number of data having $i$ and $j$ as two first BMUs.}
     \ENDFOR
\end{algorithmic}
\end{algorithm}

At the end of this step, a local density value is associated with each prototype and a connectivity value is associated with each pair of neurons. Most of the information on the data structure is summarized in these values. 

\subsection{Clustering of the prototypes}

The last step of the process is the clustering of the prototypes into a smaller number of classes.
Various  approaches have been proposed for this task \cite{Bohez98,Hussin04,Korkmaz06}. However, the obtained clustering is never optimal, since part of the information contained in the data is not represented by the prototypes. Here we use the density and connectivity to optimize the clustering (see algorithm \ref{clustalgo}). 

\begin{algorithm}[!ht]
\caption{Clustering of enriched prototypes}
\label{clustalgo}
\begin{algorithmic}[1]
 \REQUIRE the density values $D_i$ and the connectivity values $v_{i,j}$.
  \ENSURE The clusters of prototypes.
    \STATE{Extract the sets of connected neurons $P=\{C_i\}^L_{i=1}$, such as:
\[\forall m \in C_i, \exists n \in C_i \text{ such as } v_{m,n} > threshold\]}
    \STATE{In this paper $threshold = 0$.}
      \FORALL {$C_k \in P$}
          \STATE{Find the set $M(C_k)$ of density maxima.
          	\[	M(C_k)  =  \lbrace w^i\in C_k\mid D_i\geq D_j, w^j \text{ neighbor to } w^i \rbrace\]
	Prototypes $w_i$ and $w^i$ are neighbor if $v_{i,j}>threshold$.
	}
          \STATE{Determine the merging threshold matrix:
\[S = \left[ S \left(i,j\right)\right]_{i,j=1...\mid M(C_k)\mid} \]
with \[S (i,j) = \left(\frac{1}{D_i}+\frac{1}{D_j}\right)^{-1}\]}
      \FORALL {prototype $w_i \in C_k$}
             \STATE{Label $w^i$ with one element $label(i)$ of $M(C_k)$, according to an ascending density gradient along the neighborhood. Each label represents a micro-cluster.}
        \ENDFOR
      \FORALL {pair of neighbors prototypes $(w^i,w^j)$ in $C_k$}
             \STATE{merge the two micro-clusters if:
 			\[label(i) \neq label(j),\]
 			\[D_i > S(label(i),label(j)\]
 			and
 			\[D_j > S(label(i),label(j))\]}
        \ENDFOR
     \ENDFOR
\end{algorithmic}
\end{algorithm}

The main idea is that the core part of a cluster can be defined as a region with high density. Then, in most cases the cluster borders are defined either by a low density region or an ``empty'' region between clusters (i.e. large inter cluster distances) \cite{Ultsch05}.

At the end of the enrichment process, each set of prototypes with a positive connectivity values ($v$) defines well separate clusters (i.e. distance-defined). This is useful to detect borders defined by large inter-cluster distances.

The estimated  local density ($D$) allows the detection of cluster borders defined by low density. Each cluster is defined by a local maximum of density (density mode).
Then, a ``Watersheds'' method \cite{Vincent91} is applied on prototypes' density for each well separated cluster to find low density area inside these clusters, in order to characterize density defined sub-clusters.
For each pair of adjacent subgroups, we use a density-dependent index \cite{Yue04} to check if a low density area is a reliable indicator of the data structure, or whether it should be regarded as a random fluctuation in the density.
This process is very fast thanks to the small number of prototypes. The combined use of these two types of group definitions can achieve very good results despite the low number of prototypes in the map. In addition, the number of cluster is detected automatically (cf. \cite{Cabanes08a}).

\section{Experimental results}
\label{Experiments}

For testing the validity of the proposed method, we observe the performances of the proposed algorithm in comparison to other SOM-based methods that use density estimation. The differences between the compared algorithms are related to the way of computing distances between histograms. Indeed, the dissimilarity measure affects the learning of the SOM and the update of the prototypes in addition to the density computation.

We have generated six data-sets constituted by different configurations of histogram data. Each data-set points out the different characteristics of data clusters according to the different distributions parameters: mean, variation and shape.

The histogram data are generated, each one by 1000 random values, using a Gamma distribution with three parameters: the mean value, the standard deviation and a shape parameter, controlling the skewness of the distribution.
The sequence of 1000 values is shared in 10 continuous intervals corresponding to the bins of the histogram. We have chosen to build equi-depth histograms, so the bounds of the intervals correspond to the deciles of the distribution of the values. Each bin has a weight $\pi = 0.1$.

We set six data-sets according to the selected parameters:

\begin{itemize}
\item dimension $d$, with $d \in {2, 10}$;
\item number of clusters $k$, with $k \in {3, 5}$;
\item means $m$ generated by using a Normal distribution $\mathcal{N}(\mu,\sigma^2)$;
\item standard deviation $s$ generated by using a Normal distribution $\mathcal{N}(\mu,\sigma^2)$;
\item shape parameter $h$ generated by using a Normal distribution $\mathcal{N}(\mu,\sigma^2)$.
\end{itemize}

The dimension denotes the number of histograms for each observation;
the number of clusters identifies the different sub-populations;
the three parameters $m$, $s$ and $h$  of the Gamma distribution are generated by a Normal distribution $\mathcal{N}(\mu,\sigma^2)$ with $\mu$ a random value in [0,5] and $\sigma=0.1$). As standard deviations are always positive, we used a truncated Normal distribution to generate $s$ values.

For each data-set, the data are generated for different sub-populations (clusters) by a distribution that differs for one parameter ($m, s, h$) while the other two parameters have the same Normal distribution in different cluster.

The six data-sets are denoted as:

\begin{itemize}
\item DB1($d=2;k=3; \; m $)
\item DB2($d=10;k=5;\; m$)
\item DB3($d=2;k=3;\; s $)
\item DB4($d=10;k=5;\; s$)
\item DB5($d=2;k=3;\; h$)
\item DB6($d=10;k=5;\; h$)
\end{itemize}

In our proposal, the Wasserstein distance takes into account all the characteristics of the distributions.
We expect that the results are strongly depending on the distance. To validate our method (dW), we compare the results with different strategies based on different dissimilarities. We tested measures using only the component $c_i$ (dC) and $r_i$ (dR) in the Wasserstein distance decomposition (see equation \ref{HOMSQfin}). We also tested a distance (dM) based on the mean $m$ of the distributions, and a distance based on the standard deviation (dS) $s$ of the distribution. Finally, we tested two  distances between interval data computed from support values of the histograms. In the first case (dI1) has been considering the lower and upper values over the distribution support to define the interval bounds: $[min, max]$. In a second case (dI2), we considered the mean and standard deviation of the distribution to compute the interval bounds: $[\mu-\sigma, \mu+\sigma]$.

Therefore, the six approaches we compared with our proposal (dW) are based on the following distances:
\begin{itemize}
\item dC: compute the distances between data and prototypes using the difference between the centers $c_i^k$ and $c_i^l$ of each corresponding interval of the supports of the histograms.
\item dR: compute the distances between data and prototypes using the difference between the radius $r_i^k$ and $r_i^l$ of each corresponding interval of the supports of the histograms.
\item dM: compute the distances between data and prototypes using the difference between the mean values of the histograms (each histogram is represented by an unique mean value).
\item dS: compute the distances between data and prototypes using the difference between the standard deviation values of the histograms (each histogram is represented by an unique standard deviation value).
\item dI1: compute the distances between data and prototypes using the $L_2$ distance  \cite{Bock00}, each histograms being represented as an unique interval. The minimum and maximum values of each histogram is used to compute the bounds of the corresponding interval.
\item dI2: compute the distances between data and prototypes using the $L_2$ distance  \cite{Bock00}, each histograms being represented as an unique interval. Here the bounds of each interval are defined as the mean value of the corresponding histogram plus or minus the standard deviation.
\end{itemize}

The obtained results are shown in Tables \ref{ARI} to \ref{V-measure}. The performance of the different approaches is evaluated using three well known indices of quality: the adjusted Rand index \cite{Rand71}, the normalized mutual information \cite{Vinh10} and the V-measure \cite{Rosenberg07}. These indices take values in $[0,1]$, $1$ being a perfect match with the expected clustering and $0$ denoting a random solution. Internal indexes such as Silhouette, Davies-Bouldin or Dunn \cite{Arbelaitz13} are not considered in this comparisons, because they are directly based on the similarity measures that are different in the several approaches, resulting in non-comparable values despite the normalization.

\begin{table}[ht!]
\centering
\caption{Adjusted Rand Index for each data-set and each approach. $Param$ is the parameter defining the differences between the clusters, $k$ is the number of clusters, $d$ is the number of dimensions (i.e. the number of histogram per data)}
\label{ARI}
\begin{tabular}{c|c|c|c|c|c|c|c}
Data-set	&       dW&	dC&	dR&  dM&	dS& dI1& dI2\\
\hline
DB1 & 1.00	&0.88&	0.00&1.00&	0.00	&0.00	&1.00\\
DB2 & 1.00&	1.00&	0.00	&1.00&	0.00&	0.43&	1.00 \\
DB3 & 	0.95	&0.57&	0.00&	0.00&	0.00	&0.52	&0.00\\
DB4 &  0.81&	0.60	&0.00&	0.00&	0.00	&0.19	&0.00\\
DB5 &   1.00&	1.00&	0.00&	0.00&	1.00&	0.98&	1.00 \\
DB6 & 	1.00	&1.00	&0.99	&0.00	&1.00	&1.00	&1.00
\end{tabular}
\end{table}

\begin{table}[ht!]
\centering
\caption{Normalized Mutual Information for each data-set and each  approach.}
\label{NMI}
\begin{tabular}{c|c|c|c|c|c|c|c}
Data-set	&   dW&	dC&	dR&  dM&	dS& dI1& dI2\\
\hline
DB1 & 1.00 &	0.79	&0.00	&1.00&	0.00	&0.00&	1.00\\
DB2 & 1.00	&1.00&	0.00&	1.00&	0.00&	0.53&	1.00 \\
DB3 &	0.92	&0.58&	0.00&	0.00	&0.00&	0.48&	0.00\\
DB4 & 0.82&	0.66&	0.00	&0.00&	0.00&	0.37&	0.00\\
DB5 & 1.00&	1.00	&0.00	&0.00&	1.00&	0.96	&1.00 \\
DB6 & 1.00&	1.00	&0.98&	0.00&	1.00&	1.00&	1.00
\end{tabular}
\end{table}

\begin{table}[ht!]
\centering
\caption{V-measure for each data-set and each  approach.}
\label{V-measure}
\begin{tabular}{c|c|c|c|c|c|c|c}
Data-set	&       dW&	dC&	dR&  dM&	dS& dI1& dI2\\
\hline
DB1 &	1.00	&0.87&	0.00&1.00&	0.00	&0.00	&1.00\\
DB2 &	1.00&	1.00&	0.00	&1.00&	0.00&	0.63&	1.00 \\
DB3 & 0.92	&0.73&	0.00&	0.00&	0.00	&0.61	&0.00\\
DB4 & 0.84&	0.73	&0.00&	0.00&	0.00	&0.48	&0.00\\
DB5 & 1.00&	1.00&	0.00&	0.00&	1.00&	0.96&	1.00 \\
DB6 & 1.00	&1.00	&0.98	&0.00	&1.00	&1.00	&1.00
\end{tabular}
\end{table}

The combination of the two types of cluster separations turns out to be effective. Since the clusters are defined by a suitable dissimilarity between histograms, the proposed approach detects a break in the connectivity of the SOM's network. Moreover, when the clusters are overlapping, the density change (defined as the amount of similar histograms for each prototype of the SOM) defines clusters boundaries. In such away, the measure of similarity between histograms is essential to guarantee the efficiency of the approach, it is at the basis of the connectivity and  density values computation. From the result we can see that the proposed method dW, based on the $L_2$ Wasserstein,  is able to detect correctly the cluster separations (and therefore the correct number of clusters) for the 6 data-sets. This result confirms the power of the Wasserstein distance to catch all information related to the different characteristics of the distributions (mean, variability and shape). The results of the other approaches based only on some parameters, reveal their inadequacy to use the whole information concerning the distributions.

It is clear that the dM approach allows to detect clusters with different mean value, but it is not capable to separate clusters with different standard deviations or shapes. In the same way, the dS approach distinguishes only clusters characterized by different standard deviations.

The comparisons with the interval data dI1 and dI2 approaches show that, in the case of intervals defined by the mean and the standard deviation, it is possible to detect clusters with different means values or with different standard deviations, but it is not possible to well separate clusters with different shapes. Considering the min-max interval data, the approach detects clusters with different standard deviation, but it shows poor results to detect clusters with different means or shapes.

In our experiment, the comparisons with the dR approach based on the "radius" component of the Wasserstein distance decomposition are usually unable to detect the differences between clusters except for the sixth data-set. This result can be explained by the high dimension of this data-set (10 histogram variables), due to a good separation of the 5 clusters in the representation space, as well as to different standard deviations of the several clusters.

The results obtained by the dC approach based on the "center" component of the Wasserstein distance decomposition, are better than the results of the other comparing approaches (except for the proposed one). In particular, the detection of clusters with different means or standard deviations is efficient. However, clusters with different shapes are poorly detected.

Our approach is the only one able to detect correctly the differences in shape of the distributions, as well as, to detect clusters with different means or standard deviations. This approach is the combination of the center and the radius component in the $L_2$ Wasserstein distance (see equation \ref{HOMSQfin}). However, it is worth of notice that even if the approach using radius alone, it is unable to detect clusters with different shapes, the combination of this radius and the second component (the centers) improves considerably the results.

\begin{figure}[ht!]
   \center
   \includegraphics[width=0.8\linewidth]{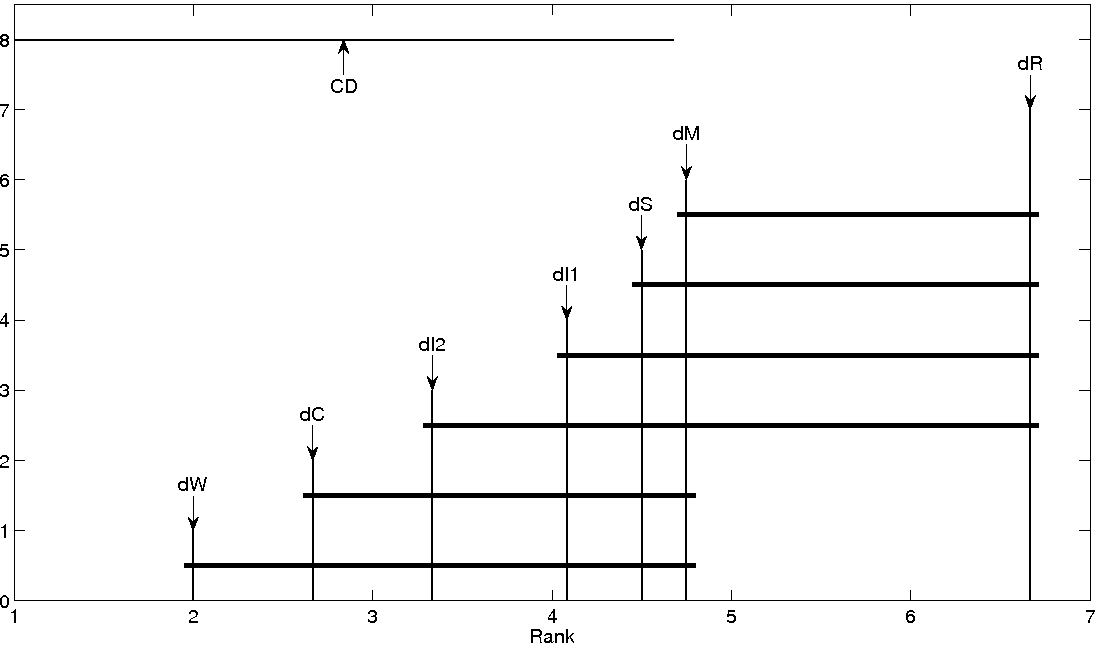}
   \caption{\label{Test} A Friedman test with post-hoc Nemenyi test on the Adj Rand index values of the compared approaches. CD is the critical distance.}
\end{figure}

In order to further validate the obtained results, a Friedman test with post-hoc Nemenyi test has been performed on the Adjusted Rand index values (table \ref{ARI}). According to the Nemenyi post-hoc test, the null hypothesis (no performance difference of two algorithms) is rejected if the average rank difference is greater than the critical distance (see \cite{Demsar2006}). In Figure \ref{Test} a critical diagram represents a projection of average ranks approaches on enumerated axis. Thick lines connecting the approaches represent the average ranks with no significant difference (for a significance threshold of $0.05$).
In this figure, the compared approaches are ordered from the best (the proposed approach, on the left) to the worst (dR, on the right).

\section{Application: China’s weather data-set}
\label{Application}
The proposed algorithm was applied to a climatic data-set, the China’s weather data-set \cite{Frank10}. Data comes from measures collected recorded monthly from 1840 to 1988 in 60 meteorological stations of the People's Republic of China. We have considered only the following variables: mean temperature, total precipitations, cloud amount,  relative humidity and wind speed. 

\begin{figure}[ht!]
   \center
   \includegraphics[width=\linewidth]{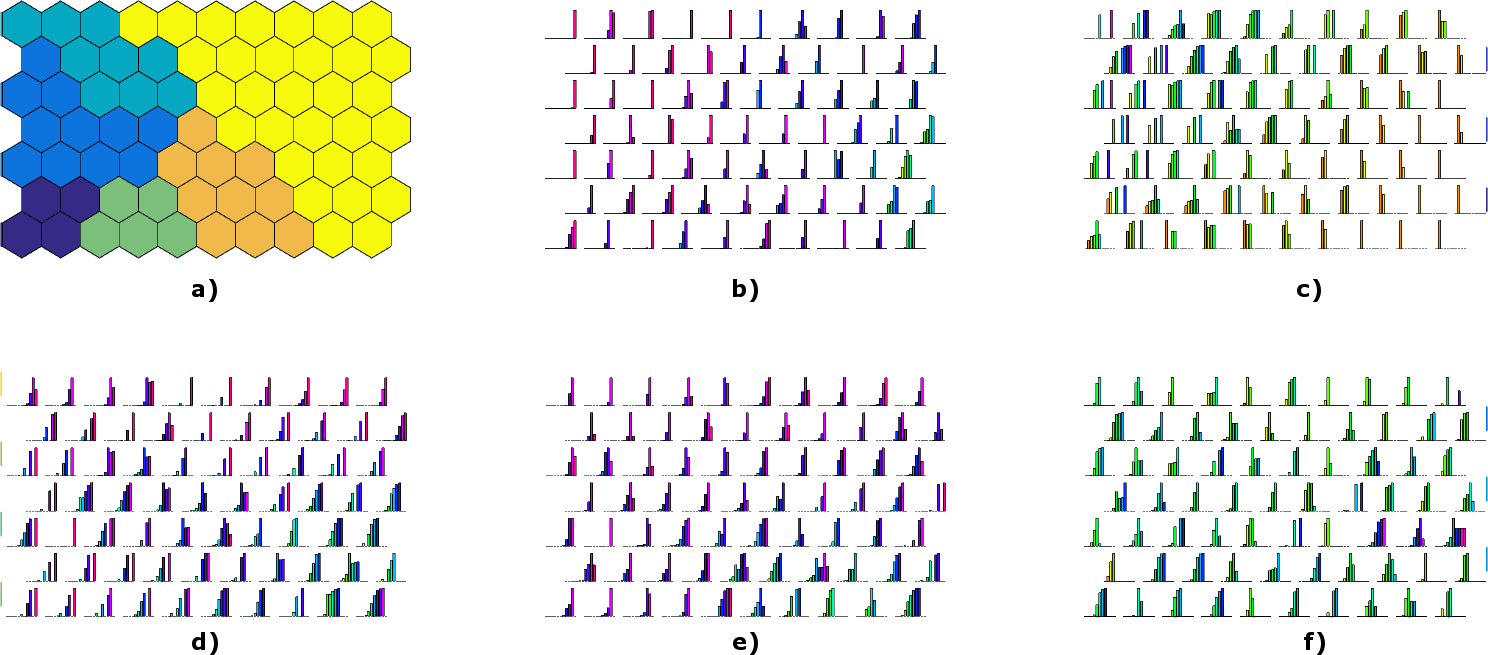}
   \caption{\label{China1} Visualization of the clustering of  the seasonal China’s weather data-set. a) The SOM is represented by an hexagonal grid where each prototype is associated to a tile. Clusters identified by the algorithm have a different color. Panels from b) to f) show the histograms of SOM prototypes for each variable (respectively mean temperature, total precipitations, cloud amount,  relative humidity and wind speed). }
\end{figure}

We have taken the distributions of these variables for each season (Summer, Fall, Winter, Spring) over the several years. The distributions of values for each season and each variable were represented by histograms and we applied the proposed algorithm to segment the data-set into several clusters. As shown in Figure \ref{China1}, the algorithm identifies six clusters. Each station is thus represented four times, one for each season. The visualization of the prototypes presented in Figure \ref{China1} (b) to (f) for each variable is a powerful tool to characterize the differences between clusters.

\begin{figure}[ht!]
   \center
   \includegraphics[width=\linewidth]{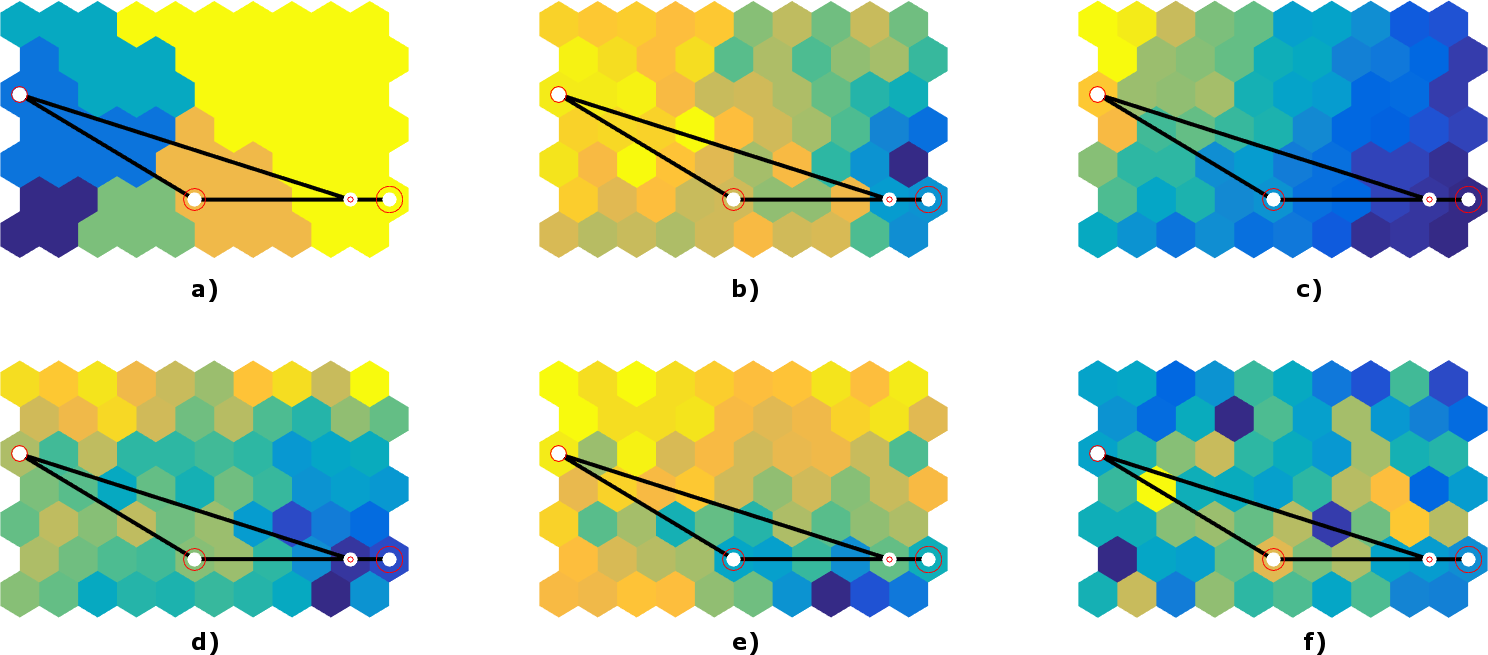}
   \caption{\label{China2} Visualization of the clustering of  the seasonal China’s weather data-set with the clusters (a) and the mean value of the prototypes for each variable (b to f). The evolution of the weather in Beijing during the year is represented as a trajectory on the maps.}
\end{figure}

Figure \ref{China2} shows the mean values (instead of the histograms) of the prototypes for the fives variables. This representation is probably easier to use, but the information about the distributions is lost. Figure \ref{China2} also shows an example of trajectory of a city (Beijing here) over the four seasons. The weather of Beijing during winter is represented in the yellow cluster (far left of the map), in spring it belongs to the orange cluster, then to the blue cluster in summer and the yellow again during fall.

\begin{figure}[ht!]
   \center
   \includegraphics[width=0.8\linewidth]{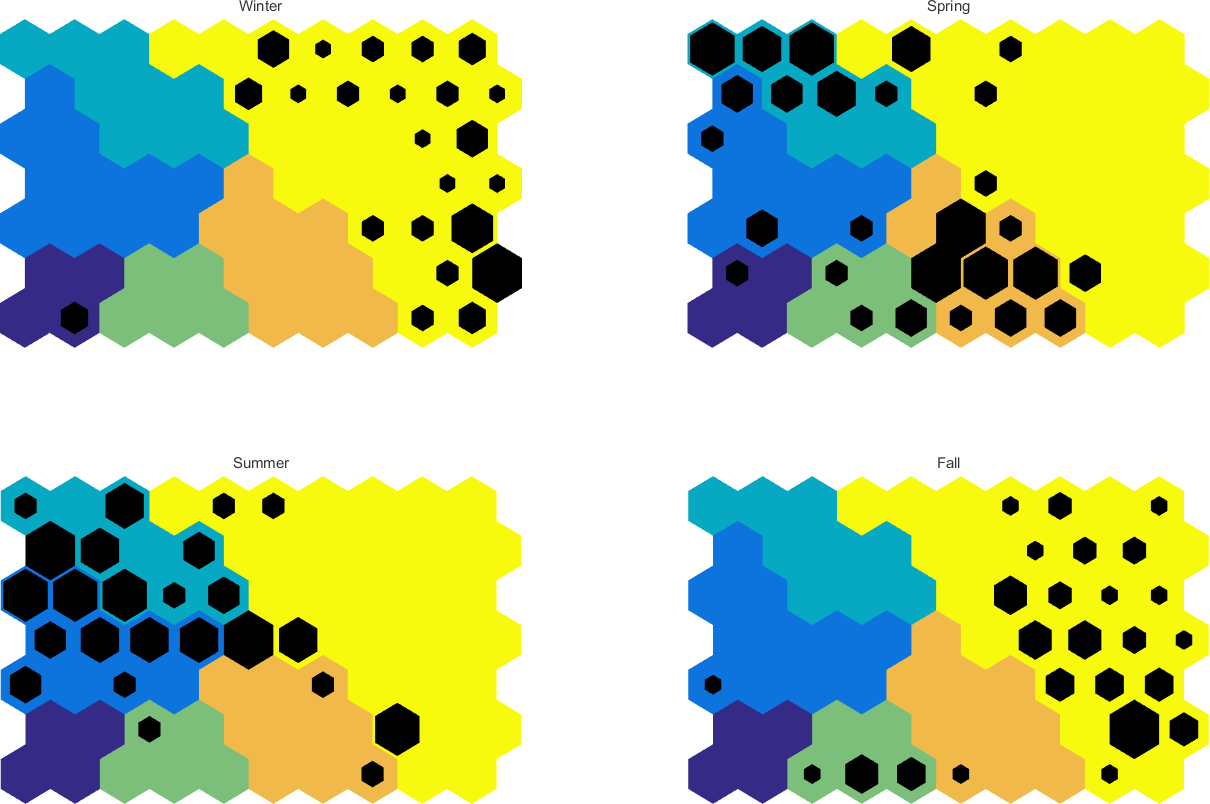}
   \caption{\label{China3} Visualization of the clustering of the seasonal China’s weather data-set. For each season, the number of cities represented by each prototype is visualized proportionally as a black hexagon.}
\end{figure}

In order to visualize the global weather variation of all of the stations during the year, it is possible to project them on the map for each season. In Figure \ref{China3}, for each season, the number of stations represented by each prototype is visualized proportionally as a black hexagon. The clustering of the map is obtained from the weather of every stations during all season, as in Figure \ref{China1} and \ref{China2}.

\begin{figure}[ht!]
   \center
   \includegraphics[width=0.8\linewidth]{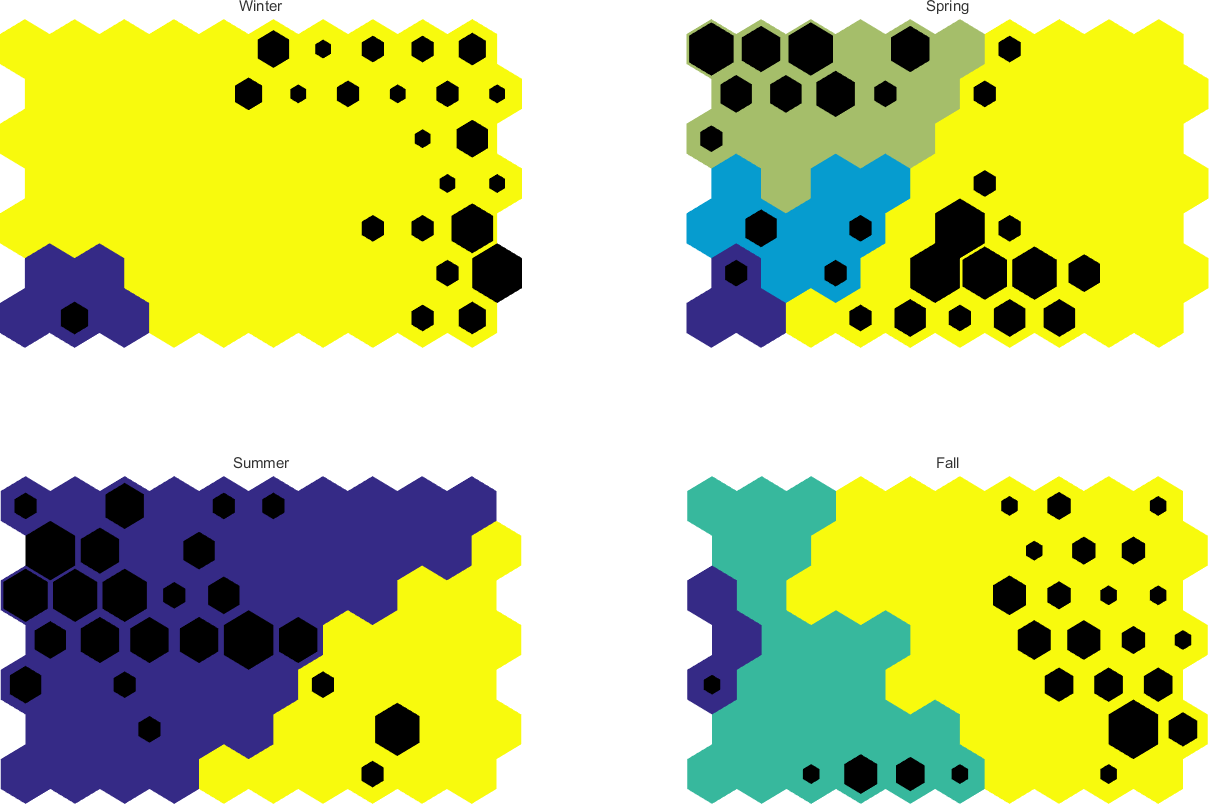}
   \caption{\label{China4} Visualization of the clustering computed for each season. The number of cities represented by each prototype is visualized proportionally by a black hexagon.}
\end{figure}

Performing a clustering for each season, it is possible to visualize the diversity of weather in each season.  In Fig. \ref{China4}, the prototypes of the map are computed for all the seasons, but the boundaries between clusters are computed only with data of one season for each sub-figure. It appears that in China, during winter, the weather of most of the stations cannot be separated in several cluster. Indeed, most stations belong to the yellow cluster during winter. This means that the weather of these stations, despite showing clear diversity (the cluster is wide on the map) changes gradually amongst them, with no abrupt change. Only few cities have a significant different weather (the blue cluster). In summer, we observe a clearer separation between two types of stations, too. However, in fall, and even more in spring, we observe a clear separation between groups of stations. Each group expresses a particular pattern of weather, with it's own variability, with few intermediate patterns.

\begin{figure}[ht!]
   \center
   \includegraphics[width=\linewidth]{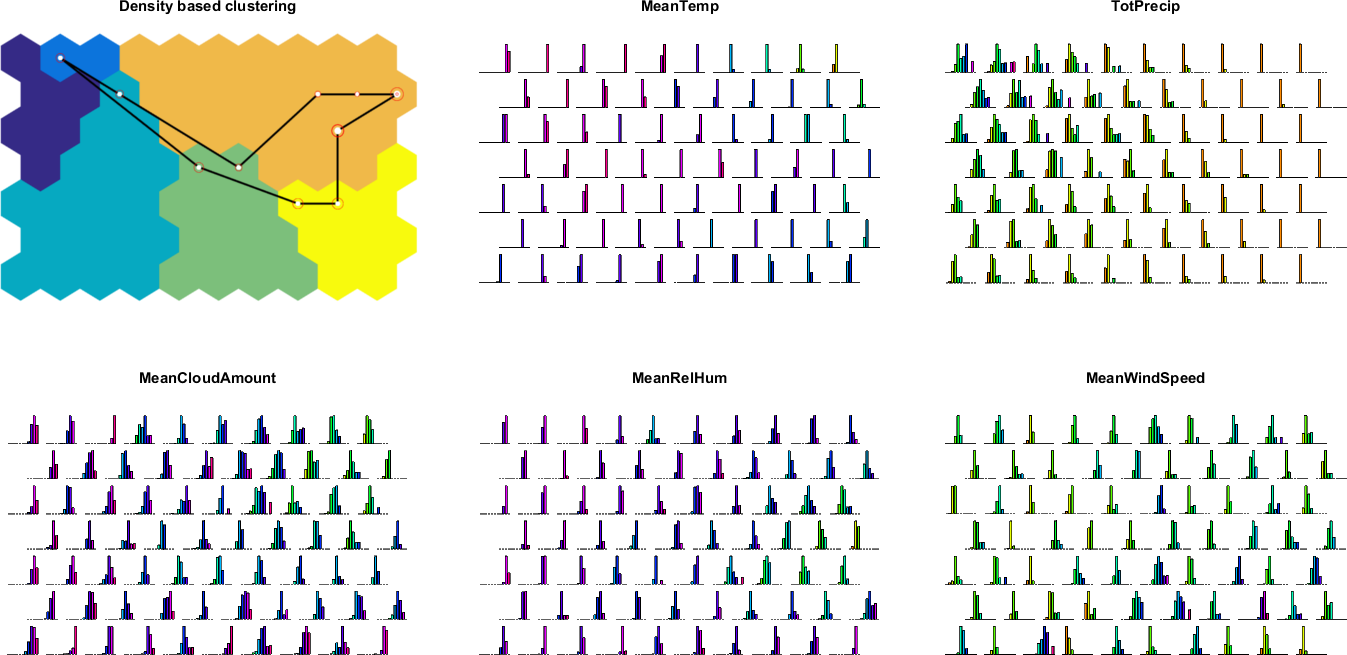}
   \caption{\label{China5} Visualization of the clustering of the monthly China’s weather data-set with the clusters and the prototypes for each variable (respectively mean temperature, total precipitations, cloud amount,  relative humidity and wind speed). The clusters are represented with different colors. The evolution of the weather in Beijing during the year is represented as a trajectory on the map.}
\end{figure}

We also considered the analysis for each month. The distributions of the monthly values were represented by histograms and we applied the proposed algorithm to segment the data-set. Each station is thus represented twelve times, one for each month, for each variable. Figure \ref{China5} shows the resulting map. The algorithm proposes again six clusters. 

Figure \ref{China5} also shows an example of trajectory of the Beijing station over the twelve months, from January (the biggest red circle on the top right part of the map) to December (the smaller circle, in the same spot as January as both months are represented by the same prototype). It is interesting to note that, during winter, the weather in Beijing is stable from October to March. All the distributions for these months are represented is the same area of the topological map. Actually, January and December are represented by the same prototype, idem with February and March. On the contrary, during summer, the climate in Beijing changes rapidly until reaching in July an highly different distribution, represented on the opposite part of the map. This is characteristic of a continental climate, typical of Beijing.

The visualization of the prototypes presented in Figure \ref{China5} for each variable allows the characterization of the different clusters. For example, there is a clear difference between the left part of the map (blue clusters) and the right part (orange and yellow) regarding the mean temperature and the total precipitation, and the distribution of cloud amount is clearly different for the orange clusters in comparison to the others.

\begin{figure}[ht!]
   \center
   \includegraphics[width=\linewidth]{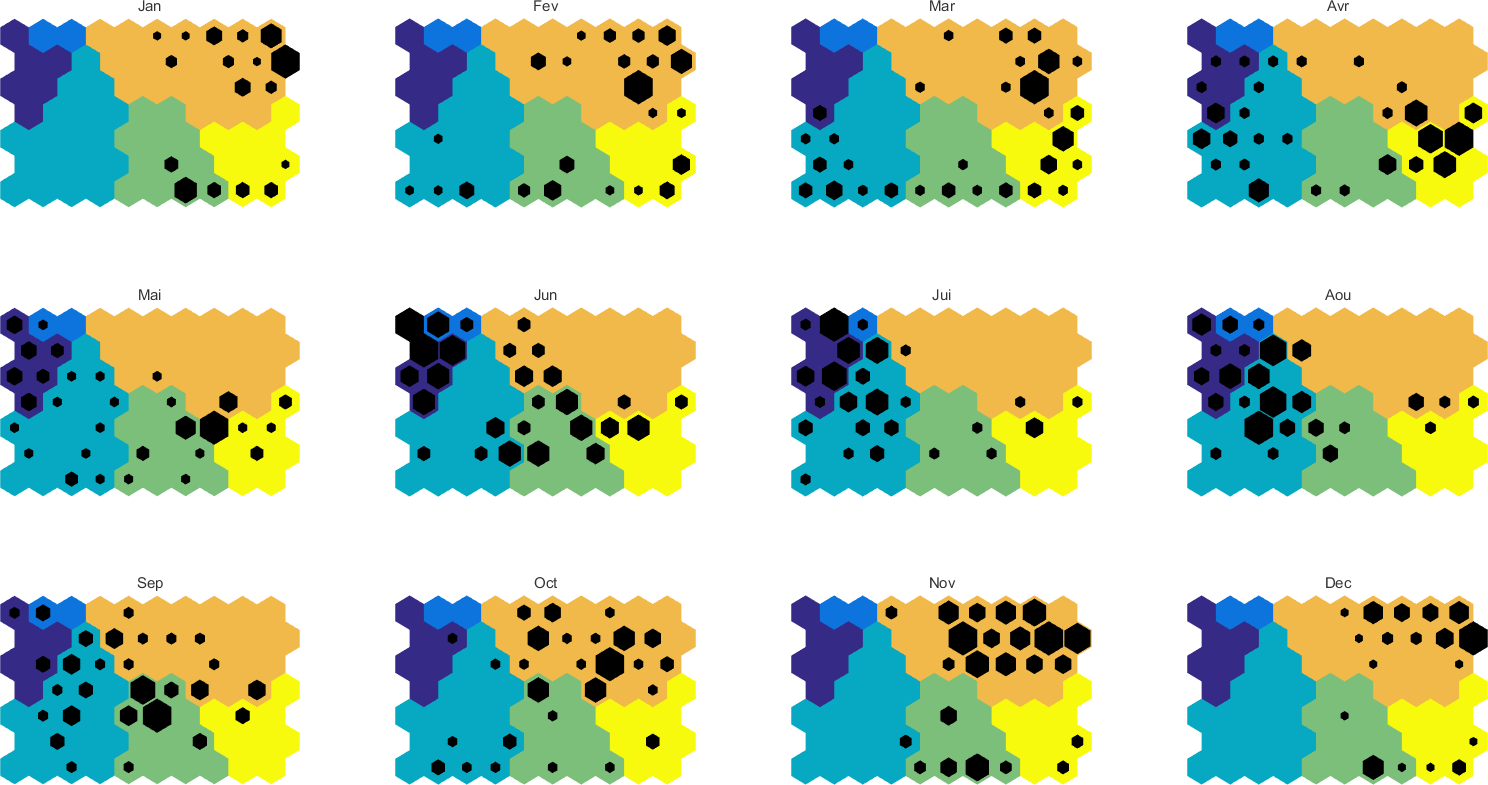}
   \caption{\label{China6} Representation of the number of cities represented by each prototype on the map for each month.}
\end{figure}

The global  weather variation in China observed monthly is represented in Figure \ref{China6}. From November to January, the stations mostly belong to the right part of the map, separated in three clusters. Then from February to June, we observe a transition from the right to the left part of the map. In July and August, many stations are in the top left area. Then, from September to November, the weather of many stations transits to the upper right corner.

\section{Conclusion}
\label{Conlu}
The clustering method presented in this paper is a two level clustering method for histogram data. The use of the $L_2$ Wasserstein metric allows to take into account all the information about position, scale and shape of the distributional data considered in the analysis. The possibility to decompose the distance in two components have shown how DS2L-SOM can point out the effect of the two distinct components of the data in partial analyses.
The proposed method is fast and doesn't require that the number of clusters is fixed in advance.
DS2L-SOM improves the classical clustering criterion by using density and connectivity measures in the partitioning process. According to the proposed method, the core part of the cluster is defined as the region with higher density. The estimation of the local density for detecting cluster borders is obtained by using the Wasserstein distances between data. The peculiarity of the presented strategy of analysis is that the density is different from the density of classical data. Considering the particular nature of the data, the used distance allows to compute the density according to the characteristics of the data distributions, enriching, so, the results. We have shown how the proposed method gives better results than concurrent strategies.
 It also allows powerful visualizations, as the application presented in section \ref{Application} shows and, in perspective, it can be extended to the analysis of the evolution of the data, following, for example, the temporal behavior of the phenomena.

\bibliographystyle{abbrv}
\bibliography{SomHist}

\end{document}